\pdfoutput=1
\documentclass[sigconf,natbib=true,nonacm=true]{acmart}

\AtBeginDocument{%
  }

\usepackage{multirow}
\usepackage{tikz}
\usepackage{algorithm}
\usepackage{algorithmic}
\usepackage{graphicx}
\usepackage{subcaption}
\usepackage{amsmath}
\usepackage{placeins}
\usepackage{capt-of}
\usepackage{cuted}
\usepackage{float}

\setcopyright{none}
\renewcommand\footnotetextcopyrightpermission[1]{}

\begin{document}

\title{Before Fusion, Ask What to Keep: Contextual Calibration of Multimodal Signals}

\author{Jiyuan Liu}
\authornote{Corresponding authors.}
\email{jiyuan.liu@student.adelaide.edu.au}
\affiliation{%
  \institution{Adelaide University}
  \city{Adelaide}
  \country{Australia}
}

\author{Liangwei Nathan Zheng}
\email{liangwei.zheng@adelaide.edu.au}
\affiliation{%
  \institution{Adelaide University}
  \city{Adelaide}
  \country{Australia}
}

\author{Wei Emma Zhang}
\email{wei.e.zhang@adelaide.edu.au}
\affiliation{%
  \institution{Adelaide University}
  \city{Adelaide}
  \country{Australia}
}

\author{Xinpei Wang}
\email{wangxinpei@sdu.edu.cn}
\affiliation{%
  \institution{Shandong University}
  \city{Jinan}
  \country{China}
}

\author{Weitong Chen}
\authornotemark[1]
\email{T.Chen@Adelaide.edu.au}
\affiliation{%
  \institution{Adelaide University}
  \city{Adelaide}
  \country{Australia}
}

\begin{abstract}
Multimodal systems often benefit from combining information across language, sound, and visual streams, but this benefit is not guaranteed. A modality that is useful for one input may become distracting for another, and local feature responses within the same modality can disagree with evidence from other sources. This work investigates how to adjust multimodal representations before they are merged by a downstream predictor. We develop a compact calibration module that compares each modality with the others at the summary level, extracts cues of cross-source support and conflict, and converts these cues into instance-wise and dimension-wise modulation signals. The calibration is applied to the original modality features rather than to already fused representations, enabling the model to suppress misleading components, preserve weak but useful evidence, and emphasize responses that are better supported by the current multimodal context. The module is designed as a plug-in component and can be attached to different fusion backbones without changing their prediction heads. Across five benchmarks covering sentiment understanding, action recognition, audio-visual event detection, and audio-visual emotion classification, the proposed pre-combination calibration strategy improves performance under both sequence-based and convolutional fusion settings. Additional analyses under modality removal, synthetic corruption, training dynamics, and feature-level visualization show that calibrating signals before fusion can reduce interference from unreliable modalities and produce more stable multimodal optimization.

\end{abstract}

\maketitle

\section{Introduction}

Multimodal learning aims to improve prediction by integrating heterogeneous signals from different modalities, such as text, vision, knowledge graph, and domain-specific signals~\cite{Zadeh17:TensorFusion,Sun25:MedicalMMSurvey,tan2026memotime,tan2026privgemo}. 
In principle, different modalities provide complementary views of the same event, and multimodal fusion has been widely used to model intra-modal and inter-modal dynamics~\cite{Zadeh17:TensorFusion,Tsai19:Mult}. 
However, adding more modalities does not always improve performance. 
Prior work shows that multimodal networks may underperform unimodal models because different modalities overfit and generalize at different rates~\cite{Wang20:GradBlending}. 
Modality imbalance can also cause dominant modalities to suppress weaker ones, leaving some unimodal representations under-optimized~\cite{Peng22:OGMGE}. 
This problem becomes more challenging when weak or less reliable modalities may introduce noisy, redundant, or conflicting information into multimodal learning~\cite{Zou23:UniSMMC}. 
Therefore, less reliable modalities may still provide useful complementary cues, but they may also contain harmful or conflicting components that interfere with full-modality prediction when directly fused into the backbone.

Existing studies mainly address this issue from an optimization-level modality imbalance perspective. 
Methods such as Grad-Blending, OGM-GE, PMR, MMPareto, and ARL improve multimodal learning by adjusting losses, gradients, modality contributions, or representation learning dynamics~\cite{Wang20:GradBlending,Peng22:OGMGE,Fan23:PMR,Wei24:MMPareto,Wei25:ARL,zhai2025graph}. 
Although effective, these methods mostly intervene after modality features have entered joint optimization. 
They therefore do not fully answer an earlier question: before multimodal fusion, \textbf{how can a model identify contextually influential but potentially unreliable modality components and decide whether they should be enhanced, retained, or suppressed?}

This question is important because direct fusion can mix useful evidence with unreliable signals before the model has an explicit chance to distinguish them. 
This motivates value estimation as a mechanism for guiding modality refinement at the pre-fusion stage.

\begin{table}[H]
\centering
\caption{Preliminary comparison between naive fusion, optimization-level balancing, and enhanced pre-fusion transformation on MOSEI.}
\label{tab:intro_preliminary}
\begin{tabular}{lcc}
\hline
Method & Acc & Macro-F1 \\
\hline
Concat & 0.8255 & 0.7810 \\
MMPareto & 0.8279 & \textbf{0.7812} \\
Concat + MLP & \textbf{0.8298} & 0.7803 \\
\hline
\end{tabular}
\end{table}

As shown in Table~\ref{tab:intro_preliminary}, simple concatenation performs poorly, while Concat+MLP reaches a performance range comparable to MMPareto. 
This suggests that lightweight pre-fusion transformation can partially mitigate modality interference before joint optimization. 
However, Concat+MLP still performs a generic feature transformation and does not explicitly estimate which modality components are contextually influential under cross-modal agreement or discrepancy. This motivates our value-aware pre-fusion refinement design.

In this paper, we define modality value as context-conditioned impact evidence for multimodal prediction. 
Rather than serving as a direct retention probability, value indicates where stronger context-aware modulation may be needed, while the final gate learns whether each component should be enhanced, retained, or reduced.

Attention-based methods mainly model token-level or cross-modal relevance~\cite{Vaswani17:Attention,Tsai19:Mult}, while generic gating mechanisms usually learn adaptive feature or modality weights~\cite{Arevalo17:GMU,Hu18:SENet}. 
Recent multimodal gating methods further explore confidence-guided gating for flexible or incomplete modality inputs~\cite{Zheng25:ConfidenceGate}. 
However, relevance, confidence, or importance does not necessarily indicate value in our setting, because a salient or confident modality can still be harmful when it conflicts with the current cross-modal context. 

Based on this definition, we propose Value-Gated Modality Refiner (VGMR), a pre-fusion refinement module for robust multimodal learning. 
VGMR first models summary-level cross-modal interactions by capturing both agreement and discrepancy between the target modality and the remaining modalities. 
It then estimates value signals at both the global and channel levels, and uses these signals as contextual guidance for a fine-grained gate generator that refines the original modality features before they enter the backbone. 
By doing so, VGMR allows the model to preserve, enhance, or reduce modality components according to both cross-modal value evidence and feature-level context before fusion.

We conduct extensive experiments and analyses to examine whether VGMR improves multimodal learning beyond simple feature aggregation, generic gating, or training-stage rebalancing. 
The results show that pre-fusion value-conditioned refinement can improve multimodal performance, enhance robustness under corrupted inputs, and support more stable optimization behaviour.

The main contributions of this paper are summarized as follows:

\begin{itemize}
    \item We revisit multimodal imbalance from a pre-fusion perspective, highlighting that weak modalities may contain both complementary cues and noisy or harmful responses.
    
    \item We propose VGMR, a value-conditioned modality refiner that estimates global-level and channel-level impact evidence and uses it to guide fine-grained gating over original modality features before fusion.
    
    \item We conduct extensive experiments and diagnostic analyses to show that VGMR improves multimodal performance, enhances robustness under noisy inputs, and supports more stable optimization behaviour.
\end{itemize}

\section{Related Work}

\subsection{Modality Imbalance and Optimization Balancing}

Modality imbalance is a common challenge in multimodal learning, where different modalities contribute unequally during joint training. 
A dominant modality may suppress weaker modalities, and adding more modalities does not always improve prediction. 
Existing methods mainly address this issue by adjusting modality contributions during training, including loss balancing, gradient modulation, modality reweighting, and representation re-learning.

Grad-Blending addresses multimodal imbalance from the perspective of overfitting and generalization differences across modalities~\cite{Wang20:GradBlending}. 
OGM-GE dynamically measures contribution discrepancy between modalities and modulates gradients to give more optimization effort to under-optimized modalities~\cite{Peng22:OGMGE}. 
PMR introduces modality prototypes to stimulate slow-learning modalities and rebalance their learning progress~\cite{Fan23:PMR}, while AGM adaptively adjusts modality gradients according to estimated modality contribution~\cite{Li23:AGM}. 
MMPareto formulates multimodal and unimodal objectives from a Pareto optimization perspective to reduce gradient conflict~\cite{Wei24:MMPareto}. 
D\&R diagnoses the learning state of each modality and re-learns modality encoders to avoid over-emphasizing scarcely informative modalities~\cite{Wei24:DR}. 
MLA reformulates joint multimodal training as alternating unimodal adaptation, reducing interference while maintaining shared prediction~\cite{Zhang24:MLA}. 
ARL further argues that perfectly balanced modality learning is not always optimal, and uses asymmetric representation learning according to modality variance and bias~\cite{Wei25:ARL}. 

Overall, these methods show that multimodal imbalance is closely related to unequal learning dynamics among modalities. 
However, they mainly intervene during joint training by adjusting losses, gradients, modality weights, or representation learning strategies. 
In contrast, this work studies whether noisy or unreliable modality response components can be refined before they enter fusion or joint optimization.

\subsection{Multimodal Fusion and Cross-Modal Interaction}

Multimodal fusion aims to combine heterogeneous modality features and exploit complementary information across modalities. 
Early fusion methods aggregate modality representations through summation, concatenation, or MLP-based fusion, while more expressive methods model higher-order cross-modal interactions. 
For example, Tensor Fusion Network explicitly captures unimodal, bimodal, and trimodal interactions~\cite{Zadeh17:TensorFusion}, and Transformer-based multimodal models use attention mechanisms to capture long-range and cross-modal dependencies~\cite{Vaswani17:Attention,Tsai19:Mult}. 
Multimodal fusion has also been widely studied in domain-specific applications, including medical multimodal learning~\cite{Sun25:MedicalMMSurvey}.

Recent studies further show that effective fusion requires more than direct aggregation. 
MISA learns modality-invariant and modality-specific subspaces to reduce modality gaps while preserving private modality characteristics~\cite{Hazarika20:MISA}. 
MAG introduces a multimodal adaptation gate to inject acoustic and visual information into pretrained language models~\cite{Rahman20:MAG}. 
MMIM improves fusion by maximizing mutual information at both the inter-modality level and the fusion-output level~\cite{Han21:MMIM}. 
Self-MM learns modality-specific representations through self-supervised unimodal label generation~\cite{Yu21:SelfMM}. 
TETFN enhances non-linguistic modalities with text-based attention and learns both consistency and differentiated information across modalities~\cite{Wang23:TETFN}. 
PCAG introduces pre-gating and contextual attention gates to filter non-informative cross-modal interactions and reduce uncertainty introduced by cross-attention~\cite{Zhang24:PCAG}.

These studies show that multimodal fusion has evolved from simple aggregation to expressive alignment, interaction, information preservation, and gating-based mechanisms. 
However, most fusion-oriented methods focus on how to combine, align, or interact modality representations. 
Even when attention or gating is used, the learned weights usually reflect relevance, salience, confidence, or feature importance. 
They do not explicitly estimate context-conditioned value evidence for refining original modality features before fusion.

\subsection{Information Decomposition Perspective}

This issue is also related to the information-decomposition perspective. 
Partial Information Decomposition (PID) provides an information-theoretic view for understanding how multiple sources contribute to a target variable by separating redundant, unique, and synergistic information~\cite{Williams10:PID}. 
This perspective is useful for multimodal learning because the contribution of a modality cannot be fully described by a single importance score. 
A modality may contain redundant information shared with other modalities, unique information from itself, or synergistic information that becomes useful only when combined with other modalities.

Our method is conceptually inspired by this view, but it does not aim to explicitly estimate PID quantities. 
Instead, PID motivates us to consider that modality components may have different contextual effects under cross-modal agreement and discrepancy. 
VGMR operationalizes this intuition through learnable global-level and channel-level value estimation, allowing useful modality-specific or complementary information to be preserved while noisy or harmful components are reduced before fusion.

\subsection{Position of This Work}

VGMR complements both optimization-level balancing methods and fusion-oriented interaction methods. Optimization-level approaches adjust losses, gradients, or modality weights during joint training, while fusion methods mainly focus on how to aggregate, align, or exchange information after modality representations have entered the fusion stage. In contrast, VGMR acts before fusion by refining original modality features with context-conditioned value evidence.

Different from attention, confidence estimation, reliability weighting, or generic gating, VGMR does not treat value as a direct preservation score or as standalone modality reliability. Instead, it estimates global-level and channel-level value signals as context-conditioned impact evidence under cross-modal agreement and discrepancy. A high value indicates that a modality or channel may strongly affect the final prediction under the current context, but it does not by itself determine whether the corresponding feature should be preserved or suppressed. The final gate combines value evidence with raw and projected feature evidence to learn the modulation direction through end-to-end task supervision.

This design separates where contextual impact may exist from how each feature should be modulated. By doing so, VGMR provides a pre-fusion value-conditioned refinement mechanism that can help preserve task-relevant complementary cues while reducing noisy or conflicting responses before multimodal fusion.

\section{Proposed Method}

\subsection{Problem Definition}

\begin{figure*}[!t]
\centering
\includegraphics[width=0.92\textwidth]{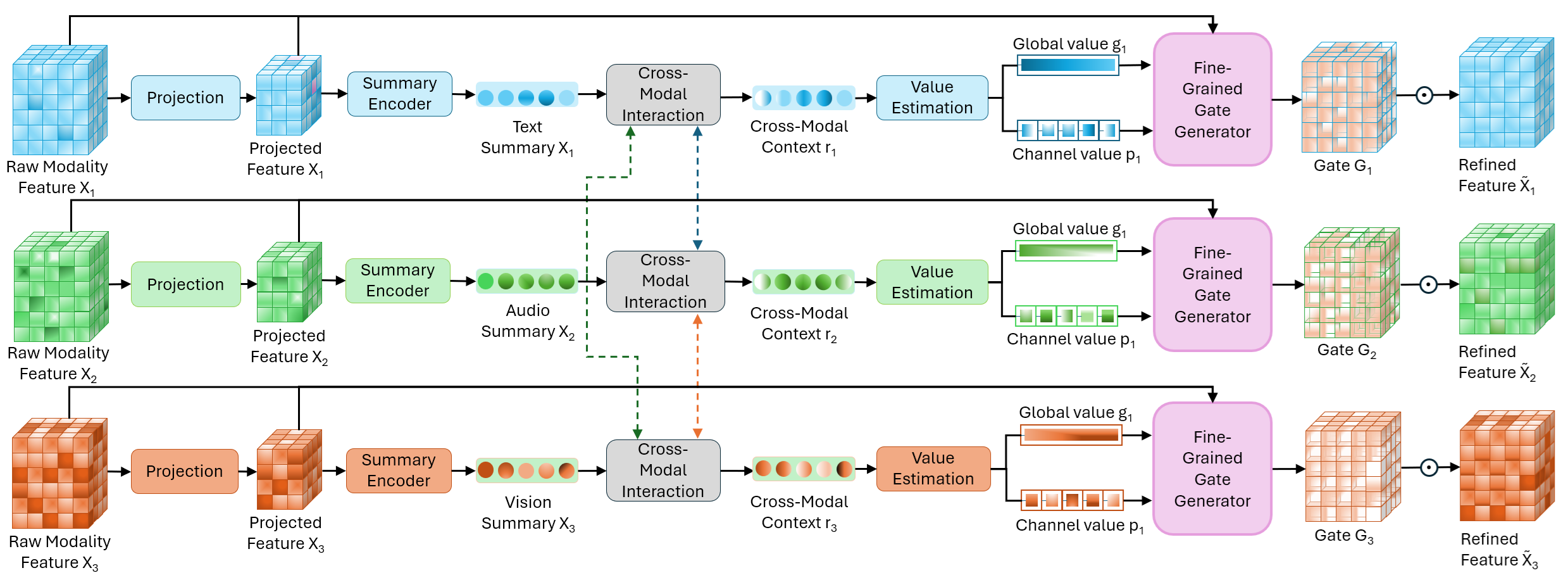}
\vspace{-1ex}
\caption{Overview of the proposed Value-Gated Modality Refiner (VGMR). 
For each modality, VGMR first projects raw features into a shared latent space and builds a summary representation (Section~3.2). 
It then constructs summary-level cross-modal interactions by modelling agreement and discrepancy between the target modality and non-target modalities (Section~3.3). 
Based on these interactions, VGMR estimates global-level and channel-level value signals as impact evidence (Section~3.4), combines them with feature evidence to generate fine-grained gates (Section~3.5), and applies gate amplification with residual retention (Section~3.6). 
The final effective gate refines the original modality feature before multimodal fusion (Section~3.7).}
\label{fig:method.png}
\vspace{-1ex}
\end{figure*}

The overall architecture of VGMR is shown in Figure~\ref{fig:method.png}. 
Given a multimodal sample with modality set $\mathcal{M}$, each modality is indexed by $m \in \mathcal{M}$. 
For text-audio-vision tasks, $\mathcal{M}=\{t,a,v\}$; for bimodal tasks, $\mathcal{M}$ contains the two available modalities, such as $\{a,v\}$.

For each modality $m$, the input feature sequence is denoted as
\begin{equation}
\mathbf{X}_m \in \mathbb{R}^{T_m \times d_m},
\end{equation}
where $T_m$ and $d_m$ are the modality-specific sequence length and feature dimension, respectively. 
VGMR aims to refine each modality before fusion by estimating context-conditioned value signals and using them to guide gate generation. 
It takes $\{\mathbf{X}_m\}_{m\in\mathcal{M}}$ as input and outputs refined features $\{\tilde{\mathbf{X}}_m\}_{m\in\mathcal{M}}$, which are then fed into the multimodal backbone.

\subsection{Modality Projection and Summary Construction}

Since different modalities have different feature dimensions, each modality is first projected into a shared latent space:
\begin{equation}
\mathbf{H}_m = P_m(\mathbf{X}_m), 
\quad \mathbf{H}_m \in \mathbb{R}^{T_m \times d_p},
\end{equation}
where $P_m(\cdot)$ is a modality-specific projection function and $d_p$ is the shared latent dimension. 
The projected feature $\mathbf{H}_m$ is used for value estimation, while the original feature $\mathbf{X}_m$ is preserved for final refinement.

We then construct a summary representation for each modality by combining average and max pooling:
\begin{equation}
\mathbf{s}_m = \phi_m\left([\operatorname{Avg}(\mathbf{H}_m);\operatorname{Max}(\mathbf{H}_m)]\right), 
\quad \mathbf{s}_m \in \mathbb{R}^{d_p},
\end{equation}
where $[\cdot;\cdot]$ denotes concatenation and $\phi_m(\cdot)$ is a modality-specific summary encoder. 
The summary vector $\mathbf{s}_m$ captures the global state of modality $m$ and is used for summary-level cross-modal interaction. 
Since interaction is performed at the summary level, VGMR does not require strict time-step alignment across modalities.

\subsection{Cross-Modal Interaction Modeling}

To estimate the value of a target modality, VGMR considers not only the modality itself but also its relationship with the remaining modalities. 
For each target modality $m \in \mathcal{M}$, we define the set of non-target modalities as
\begin{equation}
\mathcal{N}_m = \mathcal{M} \setminus \{m\}.
\end{equation}
Here, $\mathcal{N}_m$ contains all modalities except the current target modality. 
For example, in a text-audio-vision setting, when $m=t$, the non-target set is $\mathcal{N}_t=\{a,v\}$.

We first aggregate the summaries of non-target modalities to obtain a cross-modal context:
\begin{equation}
\mathbf{u}_m =
\frac{1}{\operatorname{card}(\mathcal{N}_m)}
\sum_{o \in \mathcal{N}_m} \mathbf{s}_o.
\end{equation}
Here, $\operatorname{card}(\mathcal{N}_m)$ denotes the number of non-target modalities, and $\mathbf{u}_m \in \mathbb{R}^{d_p}$ represents the average contextual reference provided by the remaining modalities. 
In a bimodal setting, $\mathbf{u}_m$ is simply the summary of the other modality; in a trimodal setting, it is the average summary of the two non-target modalities.

For each non-target modality $o \in \mathcal{N}_m$, we compute pairwise agreement and discrepancy:
\begin{equation}
\mathbf{a}_{m,o}=\mathbf{s}_m \odot \mathbf{s}_o,
\qquad
\boldsymbol{\delta}_{m,o}=|\mathbf{s}_m-\mathbf{s}_o|.
\end{equation}
We denote the collections of pairwise agreement and discrepancy terms as
\begin{equation}
\mathcal{A}_m=\{\mathbf{a}_{m,o}\mid o\in\mathcal{N}_m\},
\qquad
\mathcal{D}_m=\{\boldsymbol{\delta}_{m,o}\mid o\in\mathcal{N}_m\}.
\end{equation}
The interaction input is then constructed by concatenating the target summary, non-target summaries, pairwise agreement terms, pairwise discrepancy terms, and the averaged context:
\begin{equation}
\mathbf{r}_m =
\psi_m\left(
[\mathbf{s}_m;
\{\mathbf{s}_o\}_{o\in\mathcal{N}_m};
\{\mathbf{a}_{m,o}\}_{o\in\mathcal{N}_m};
\{\boldsymbol{\delta}_{m,o}\}_{o\in\mathcal{N}_m};
\mathbf{u}_m]
\right).
\end{equation}
The agreement terms provide feature-wise co-activation evidence between the target modality and each non-target modality, while the discrepancy terms provide pairwise mismatch evidence.
It captures feature-wise co-activation and indicates which latent dimensions are supported by both the target modality and the remaining modalities. 
The discrepancy term $\boldsymbol{\delta}_m \in \mathbb{R}^{d_p}$ is computed by the absolute difference between summaries. 
It captures cross-modal mismatch, which may reflect complementary information, modality conflict, or potential noise.
Thus, the pairwise agreement terms provide consistency evidence, while the pairwise discrepancy terms provide mismatch evidence.

\subsection{Modality Value Estimation}

Given the modality summary $\mathbf{s}_m$ and the cross-modal interaction representation $\mathbf{r}_m$, VGMR first learns a hidden value representation:
\begin{equation}
\mathbf{z}_m = \eta_m([\mathbf{s}_m;\mathbf{r}_m]),
\quad \mathbf{z}_m \in \mathbb{R}^{d_p}.
\end{equation}
Here, $\eta_m(\cdot)$ is a learnable modality-specific value encoder, and $\mathbf{z}_m$ is the latent representation used for value estimation.
The input $[\mathbf{s}_m;\mathbf{r}_m]$ combines modality-specific information with cross-modal context, agreement, and discrepancy. 
Therefore, $\mathbf{z}_m$ should be understood as context-conditioned impact evidence, rather than as a direct useful/useless classifier.

Based on $\mathbf{z}_m$, VGMR estimates a global value signal and a channel-level value signal:
\begin{equation}
g_m = \sigma((\mathbf{w}_m^g)^\top \mathbf{z}_m + b_m^g), 
\qquad
\mathbf{p}_m = \sigma(\mathbf{W}_m^p \mathbf{z}_m + \mathbf{b}_m^p),
\end{equation}
where $g_m \in \mathbb{R}$ and $\mathbf{p}_m \in \mathbb{R}^{d_p}$. 
The sigmoid function $\sigma(\cdot)$ maps the estimated value signals into a bounded range. 
The vector $\mathbf{w}_m^g \in \mathbb{R}^{d_p}$ and scalar bias $b_m^g \in \mathbb{R}$ are modality-specific parameters used to produce the global value $g_m$. 
This scalar captures sample-level impact evidence for the whole modality under the current cross-modal context. 

The matrix $\mathbf{W}_m^p \in \mathbb{R}^{d_p \times d_p}$ and bias vector $\mathbf{b}_m^p \in \mathbb{R}^{d_p}$ are used to produce the channel-level value $\mathbf{p}_m$. 
Unlike $g_m$, which gives an overall value estimate for modality $m$, $\mathbf{p}_m$ provides dimension-wise value evidence in the shared latent space. 
This allows VGMR to identify which latent channels may require stronger modulation.

The global value and channel-level value play different roles. 
The global value $g_m$ provides coarse modality-level context, while the channel value $\mathbf{p}_m$ provides finer dimension-level evidence. 
Both are used to condition gate generation, but neither of them is a direct retention mask. 
A larger value means that the corresponding modality or channel has stronger contextual impact, which may lead to enhancement or suppression depending on the final gate generator.

\subsection{Fine-Grained Gate Generation}

Before being mapped into gate logits, the bounded value signals are centered:
\[
g'_m = 2g_m - 1, \qquad \mathbf{p}'_m = 2\mathbf{p}_m - 1.
\]

\begin{equation}
\mathbf{G}_m =
\sigma\big(
\rho_m^x(\mathbf{X}_m)
\oplus \rho_m^h(\mathbf{H}_m)
\oplus \rho_m^g(g'_m)
\oplus \rho_m^p(\mathbf{p}'_m)
\big).
\end{equation}

Here, $\oplus$ denotes element-wise addition after shape alignment. 
Each mapping function $\rho(\cdot)$ transforms or broadcasts its input to the same shape as the original feature, namely $\mathbb{R}^{T_m \times d_m}$.

Specifically, $\rho_m^x(\cdot)$ is a dimension-preserving affine layer applied to the original feature $\mathbf{X}_m$. 
It provides raw feature evidence and allows the gate to depend directly on the original modality representation. 
$\rho_m^h(\cdot)$ maps the projected feature $\mathbf{H}_m$ from the shared latent dimension $d_p$ back to the original feature dimension $d_m$. 
It introduces latent-space information learned during value estimation into gate generation.

The mapping $\rho_m^g(\cdot)$ transforms the centered global value $g'_m$ into a $d_m$-dimensional vector and broadcasts it along the temporal dimension $T_m$. 
This provides sample-level modality context to every time step and feature dimension of the gate. 
The mapping $\rho_m^p(\cdot)$ transforms the centered channel-level value $\mathbf{p}'_m$ from $d_p$ to $d_m$ and also broadcasts it along $T_m$. 
This provides dimension-level value evidence for fine-grained feature modulation.

The encoders $\phi_m$, $\psi_m$, and $\eta_m$ are implemented as lightweight two-layer MLPs with nonlinear activation and output dimension $d_p$.
The functions $\phi_m$, $\psi_m$, $\rho_m^x$, $\rho_m^h$, $\rho_m^g$, and $\rho_m^p$ are modality-specific because different modalities may have different feature distributions and dimensions. 
The value encoders follow the same architecture across modalities, while their parameters are modality-specific to account for heterogeneous feature distributions. Thus, the hidden value representations share a consistent functional definition without forcing parameter sharing across different modalities.

The resulting gate $\mathbf{G}_m$ has the same shape as $\mathbf{X}_m$, enabling time-step-level and feature-dimension-level modulation. 
Unlike a generic feature gate, $\mathbf{G}_m$ is conditioned on both feature evidence and cross-modal value evidence. 
Therefore, VGMR does not assume that high value means preservation or that low value means suppression. 
Instead, value signals indicate where contextual impact may exist, while the final gate learns whether each component should be enhanced, retained, or reduced.

\subsection{Gate Amplification and Residual Retention}

Since the sigmoid gate $\mathbf{G}_m$ is limited to $[0,1]$, directly using it may lead to conservative modulation. 
We therefore transform it into an amplification gate centered around $1$:
\begin{equation}
\hat{\mathbf{G}}_m = 1 + \alpha(2\mathbf{G}_m - 1),
\end{equation}
where $\alpha \in [0,1]$ controls the amplification strength. 
When $\alpha=0$, the gate becomes neutral and does not change the feature scale. 
A larger $\alpha$ allows stronger enhancement when $\mathbf{G}_m>0.5$ and stronger suppression when $\mathbf{G}_m<0.5$.

To avoid over-suppression, we introduce residual retention:
\begin{equation}
\bar{\mathbf{G}}_m = \beta + (1-\beta)\hat{\mathbf{G}}_m,
\end{equation}
where $\beta \in [0,1]$ controls the retention strength. 
A larger $\beta$ preserves more original information and reduces the risk of discarding weak but useful modality cues. 
A smaller $\beta$ allows stronger adaptive refinement. 
This design allows the final effective gate $\bar{\mathbf{G}}_m$ to balance \textbf{adaptive refinement} and \textbf{information preservation}.

\subsection{Modality Refinement}

Finally, the effective gate $\bar{\mathbf{G}}_m$ is applied to the original modality feature:
\[
\tilde{\mathbf{X}}_m =
\operatorname{Post}_m(\mathbf{X}_m \odot \bar{\mathbf{G}}_m),
\quad
\tilde{\mathbf{X}}_m \in \mathbb{R}^{T_m \times d_m}.
\]

Here, $\operatorname{Post}_m(\cdot)$ is a lightweight dimension-preserving layer implemented as LayerNorm followed by Dropout. 
The refined feature $\tilde{\mathbf{X}}_m$ keeps the same shape as $\mathbf{X}_m$, while its components are modulated using both value evidence and feature-level information before being fed into the multimodal backbone.

\subsection{Training Objective}

The refined modality features are fed into a multimodal backbone to produce the final prediction:
\[
\hat{y}=f(\{\tilde{\mathbf{X}}_m\}_{m\in\mathcal{M}}).
\]
The whole model is trained end-to-end with:
\[
\mathcal{L}
=
\mathcal{L}_{task}(\hat{y},y)
+
\lambda_v\mathcal{L}_{value}
+
\lambda_r\mathcal{L}_{reg}.
\]
Here, $\mathcal{L}_{task}$ is the standard prediction loss, which provides the main supervision for both the backbone and the gate generator. 
The value supervision term $\mathcal{L}_{value}$ encourages the estimated value signals to reflect empirical modality contribution. 
For each modality, we estimate this contribution by the loss increase after removing it:
\[
\Delta_m
=
\ell(f(\{\tilde{\mathbf{X}}_o\}_{o\in\mathcal{M}\setminus\{m\}}),y)
-
\ell(f(\{\tilde{\mathbf{X}}_o\}_{o\in\mathcal{M}}),y).
\]
We align the global value $g_m$ and the average channel value $\operatorname{Avg}(\mathbf{p}_m)$ with $\Delta_m$, so that value scores serve as context-conditioned impact evidence rather than arbitrary gate-internal activations. 
The regularization term $\mathcal{L}_{reg}$ stabilizes the value/gate distributions and prevents value scores from collapsing to constants.

The value terms do not directly decide whether a feature should be preserved or suppressed. 
Instead, they provide contextual impact evidence, while the final gate learns the modulation direction through end-to-end task supervision.

\section{Experiments}

\subsection{Experimental Setup}

\label{sec:dataset}

We evaluate VGMR on five multimodal datasets covering sentiment analysis, action recognition, and audio-visual recognition. MOSI~\cite{Zadeh16:MOSI} and MOSEI~\cite{Zadeh18:MOSEI} are text-audio-visual sentiment analysis benchmarks. MOSI contains 2,199 utterance-level video clips from 93 opinion videos, with sentiment scores ranging from -3 to 3, while MOSEI is a larger-scale benchmark for utterance-level multimodal sentiment prediction. For both datasets, we use the binary positive/negative classification setting.

UCF101~\cite{Soomro12:UCF101} is used for human action recognition and contains 101 action classes, with 9,537 videos for training and 3,783 videos for testing under the standard protocol. Each UCF101 video is represented using RGB frames and optical flow sequences, where RGB captures appearance information and optical flow captures motion information. AVE~\cite{Tian18:AVE} is an audio-visual event dataset with 4,143 10-second videos across 28 event categories. We extract visual frames and corresponding audio clips from event-localized segments and follow the original split. CREMA-D~\cite{Cao14:CREMAD} is an audio-visual emotion recognition dataset with 7,442 clips from six emotion categories, split into 6,698 training samples and 744 testing samples.

\begin{strip}
\centering
\vspace{-8pt}
\captionof{table}{Performance comparison of different methods under concatenation and Transformer backbone settings across multiple multimodal datasets.}
\label{tab:concat_transformer_results_all}
\resizebox{\textwidth}{!}{
\begin{tabular}{c|cccccccccc|cccc}
\hline
Backbone 
& \multicolumn{10}{c|}{Concatenation} 
& \multicolumn{4}{c}{Transformer} \\
\hline
Dataset 
& \multicolumn{2}{c}{MOSI} 
& \multicolumn{2}{c}{MOSEI} 
& \multicolumn{2}{c}{UCF101} 
& \multicolumn{2}{c}{AVE} 
& \multicolumn{2}{c|}{CREMA-D} 
& \multicolumn{2}{c}{MOSI} 
& \multicolumn{2}{c}{MOSEI} \\
\hline
Method 
& Acc & F1 
& Acc & F1 
& Acc & F1 
& Acc & F1 
& Acc & F1 
& Acc & F1 
& Acc & F1 \\
\hline
Baseline
& 0.7507 & 0.7507 
& 0.8154 & 0.7850 
& 0.7864 & 0.7811 
& 0.6169 & 0.5816 
& 0.5054 & 0.5030 
& 0.7741 & 0.7732 
& 0.8255 & 0.7810 \\
MLA 
& 0.7726 & 0.7712 
& 0.8304 & 0.7893 
& 0.7803 & 0.7747 
& 0.6318 & 0.5892 
& 0.5108 & 0.5081 
& 0.7813 & 0.7801 
& 0.8184 & 0.7737 \\
D\&R 
& 0.7828 & 0.7820 
& 0.8221 & \textbf{0.7904} 
& 0.7906 & 0.7834 
& 0.5995 & 0.5614 
& 0.4852 & 0.4678 
& 0.8017 & 0.7990
& 0.8311 & 0.7848 \\
OGM-GE 
& 0.7930 & \textbf{0.7926} 
& 0.8234 & 0.7707 
& 0.7906 & 0.7849 
& 0.6468 & 0.6133 
& 0.5148 & 0.5096 
& 0.7843 & 0.7790 
& 0.8264 & 0.7724 \\
ARL 
& 0.7886 & 0.7869 
& 0.8225 & 0.7881 
& 0.7917 & 0.7832 
& 0.6567 & \textbf{0.6224}
& 0.5753 & \textbf{0.5794} 
& 0.7843 & 0.7819 
& 0.8268 & 0.7856 \\
Grad-Blending 
& 0.7857 & 0.7804 
& 0.8251 & 0.7679 
& 0.7930 & 0.7859 
& 0.6567 & 0.6109 
& 0.5175 & 0.5165 
& 0.7886 & 0.7818 
& 0.8337 & 0.7921 \\
MMPareto 
& 0.7843 & 0.7839 
& 0.8281 & 0.7760 
& 0.7909 & 0.7836 
& 0.6443 & 0.6030 
& 0.4973 & 0.4963 
& 0.7901 & 0.7839 
& 0.8279 & 0.7812 \\
PMR 
& 0.7843 & 0.7831 
& 0.8229 & 0.7696 
& 0.7975 & 0.7896 
& 0.6418 & 0.5990 
& 0.5175 & 0.5155 
& 0.7901 & 0.7884 
& 0.8332 & 0.7820 \\
AGM 
& 0.7828 & 0.7804 
& 0.8300 & 0.7899 
& 0.8020 & 0.7941 
& 0.6294 & 0.5955 
& 0.4960 & 0.4934 
& 0.8003 & 0.7907 
& 0.8302 & 0.7940 \\
\textbf{VGMR} 
& \textbf{0.7945} & 0.7911 
& \textbf{0.8356} & 0.7894 
& \textbf{0.8126} & \textbf{0.8059} 
& \textbf{0.6592} & 0.6199
& \textbf{0.5780} & 0.5782 
& \textbf{0.8076} & \textbf{0.8019} 
& \textbf{0.8446} & \textbf{0.8029} \\
\hline
\end{tabular}
}
\vspace{-8pt}
\end{strip}

All methods follow the same data splits and are evaluated using Accuracy and Macro-F1. Unless otherwise stated, we set the gate amplification coefficient $\alpha$ to 0.5 and the residual retention coefficient $\beta$ to 0.2 across all experiments.
For reproducibility, all compared methods use the same preprocessing, backbone, optimizer, batch size, stopping criterion, and random seeds, with the best validation checkpoint used for test evaluation.

For AVE and CREMA-D, the ResNet+Concat setting uses fixed lightweight audio-visual features and a shallow fusion head rather than a large pretrained video or audio foundation backbone. 
Therefore, the absolute CREMA-D numbers should be interpreted as results under this controlled low-capacity protocol, whose purpose is to test whether VGMR improves the same backbone, rather than as state-of-the-art CREMA-D performance.

\subsection{Overall Performance}

Table~\ref{tab:concat_transformer_results_all} reports the overall performance across five datasets and two backbone settings. We report Accuracy and Macro-F1 for all methods.

As shown in Table~\ref{tab:concat_transformer_results_all}, VGMR achieves the best Accuracy in most settings and obtains strong or competitive Macro-F1 across sentiment analysis, action recognition, and audio-visual recognition tasks. 

Under the ResNet+Concat backbone, VGMR consistently improves over the plain baseline on MOSI, MOSEI, UCF101, AVE, and CREMA-D, showing that the proposed refiner is not limited to text-audio-visual sentiment datasets. Under the Transformer backbone, VGMR achieves the best results on both MOSI and MOSEI, reaching 0.8446 Accuracy and 0.8029 Macro-F1 on MOSEI. Although ARL obtains slightly higher Macro-F1 on AVE and CREMA-D, VGMR achieves the best Accuracy and remains competitive in Macro-F1. Overall, these results suggest that VGMR can serve as a general pre-fusion refinement module across different backbones and tasks.

\subsection{Component Ablation}
We conduct two groups of ablation studies on MOSEI. 
The first group studies the value signals used for gate conditioning, including the global value $g_m$ and the channel-level value $\mathbf{p}_m$.

The global value $g_m$ provides sample-level modality impact evidence, while the channel value $\mathbf{p}_m$ provides dimension-level evidence for fine-grained feature modulation. 
The second group studies the cross-modal evidence used for value estimation, including the agreement term $\mathbf{a}_m$ and the discrepancy term $\boldsymbol{\delta}_m$. 
Here, $\mathbf{a}_m$ captures cross-modal consistency through feature-wise co-activation, while $\boldsymbol{\delta}_m$ captures cross-modal mismatch or potential conflict.

\begin{center}
\captionof{table}{Component ablation of VGMR on MOSEI.}
\label{tab:ablation_components}

\begingroup
\small
\setlength{\tabcolsep}{6pt}
\renewcommand{\arraystretch}{1.15}

\begin{tabular}{lcc}
\hline
Condition & Acc & Macro-F1 \\
\hline
Full VGMR & \textbf{0.8446} & \textbf{0.8029} \\
\hline
\multicolumn{3}{l}{\textit{Value signal ablation}} \\
w/o channel-signal $\mathbf{p}_m$ & 0.8324 & 0.7831 \\
w/o global-signal $g_m$ & 0.8390 & 0.8011 \\
w/o $g_m$ and $\mathbf{p}_m$ & 0.8287 & 0.7917 \\
\hline
\multicolumn{3}{l}{\textit{Agreement--discrepancy ablation}} \\
w/o agreement $\mathcal{A}_m$ & 0.8347 & 0.7965 \\
w/o discrepancy $\mathcal{D}_m$ & 0.8330 & 0.7908 \\
w/o both $\mathcal{A}_m$ and $\mathcal{D}_m$ & 0.8257 & 0.7812 \\
\hline
\end{tabular}

\endgroup
\end{center}

Table~\ref{tab:ablation_components} shows that each component contributes to the final performance of VGMR. 
Removing both global and channel-level value signals leads to a clear performance drop, indicating that explicit value evidence is important for gate generation. 
Using only $g_m$ gives limited improvement, suggesting that a single sample-level value signal is too coarse for fine-grained modality refinement. 
Using only $\mathbf{p}_m$ performs better, especially on Macro-F1, which shows that channel-level value evidence is more directly useful for selective feature modulation. 
However, the full model still achieves the best result, indicating that $g_m$ and $\mathbf{p}_m$ provide complementary conditioning: the former supplies overall modality-level context, while the latter supports finer channel-level refinement.

The agreement--discrepancy ablation further shows that value estimation benefits from both consistency and conflict evidence. 
Removing either $\mathbf{a}_m$ or $\boldsymbol{\delta}_m$ weakens performance, and removing both causes the largest drop. 
This suggests that agreement helps identify modality components supported by other modalities, while discrepancy helps detect unreliable or conflicting responses. 
The larger decrease caused by removing $\boldsymbol{\delta}_m$ indicates that conflict evidence is particularly important for weak-modality refinement. 
Overall, the best performance is obtained only when global value, channel value, agreement, and discrepancy are used together, confirming that VGMR relies on multi-level value evidence rather than a single scalar score or a generic feature gate.

\subsection{Value-Guided Gate Behavior Analysis}

We conduct a text-corruption analysis on MOSI with the Concat backbone. 
In the corrupted condition, 30\% of text timesteps are replaced with Gaussian noise. 
We focus on the text modality and report four statistics: global value, channel value, channel-value standard deviation, and average initial gate response. 
The global value reflects sample-level contextual impact, while the channel value reflects dimension-level impact evidence. 
The channel-value standard deviation measures how unevenly the value evidence is distributed across channels. 
The gate response indicates the average initial modulation strength assigned to the corrupted or clean text features.

For Global Value, Channel Value, and Gate, results are reported as mean $\pm$ sample standard deviation across test samples. 
For Channel Std, we first compute the standard deviation across channels within each sample, and then report its mean and standard deviation across the test set.

\begin{center}
\captionof{table}{Value and gate response under text corruption on MOSI.}
\label{tab:value_gate_response}

\begingroup
\small
\setlength{\tabcolsep}{3pt}
\renewcommand{\arraystretch}{1.12}

\begin{tabular}{lcccc}
\hline
Condition & Global Value & Channel Value & Channel Std & Gate \\
\hline
Clean Text 
& 0.598 $\pm$ 0.119 
& 0.554 $\pm$ 0.092 
& 0.102 $\pm$ 0.120 
& 0.481 $\pm$ 0.039 \\
Noisy Text 
& 0.788 $\pm$ 0.062 
& 0.707 $\pm$ 0.030 
& 0.301 $\pm$ 0.053 
& 0.417 $\pm$ 0.009 \\
$\Delta$ 
& +0.190 
& +0.153 
& +0.199 
& -0.064 \\
\hline
\end{tabular}

\endgroup
\end{center}

As shown in Table~\ref{tab:value_gate_response}, text corruption increases the global value, channel value, and channel-value standard deviation, but decreases the average gate response. 
This pattern suggests that the corrupted text produces stronger contextual impact and more uneven channel-level evidence, indicating that the model detects it as a high-impact input requiring modulation. 
However, the lower gate response shows that higher value does not directly mean stronger preservation. 
Instead, value signals provide impact evidence, and the gate generator combines this evidence with feature information to suppress unreliable high-impact responses before fusion.

\subsection{Controlled Comparison and Efficiency Analysis}

\begin{table}[H]
\centering
\caption{Fusion strategy comparison on MOSEI with the Transformer backbone.}
\label{tab:fusion_mosei_transformer}
\small
\setlength{\tabcolsep}{6pt}
\renewcommand{\arraystretch}{1.15}
\begin{tabular}{lcc}
\hline
Method & Acc & Macro-F1 \\
\hline
Summation & 0.8219 & 0.7785 \\
Tensor Fusion & 0.8289 & 0.7871 \\
Concat+MLP & 0.8298 & 0.7803 \\
\textbf{VGMR} & \textbf{0.8446} & \textbf{0.8029} \\
\hline
\end{tabular}
\end{table}

To examine whether the improvement of VGMR mainly comes from stronger feature aggregation, generic gating, or simply a larger parameter budget, we compare it with early-fusion strategies and pre-fusion interaction/gating mechanisms on MOSEI with the Transformer backbone. 
All controlled baselines use the same input features, data split, Transformer backbone, and training protocol. 

\begin{table}[H]
\centering
\caption{Comparison with generic interaction and gating mechanisms on MOSEI with the Transformer backbone.}
\label{tab:mechanism_comparison_mosei}
\small
\setlength{\tabcolsep}{6pt}
\renewcommand{\arraystretch}{1.15}
\begin{tabular}{lcc}
\hline
Method & Acc & Macro-F1 \\
\hline
Concat & 0.8255 & 	0.7810 \\
Cross-Attention & 0.8317 & 0.7941 \\
Sigmoid+Tanh Gate & 0.8326 & 0.7924 \\
\textbf{VGMR} & \textbf{0.8446} & \textbf{0.8029} \\
\hline
\end{tabular}
\end{table}

\begin{table}[H]
\centering
\caption{Efficiency comparison on MOSEI with the Transformer backbone.}
\label{tab:efficiency_mosei}
\footnotesize
\setlength{\tabcolsep}{2.8pt}
\renewcommand{\arraystretch}{1.05}
\begin{tabular}{lcccc}
\hline
Method & Params & Extra Params & Train/Epoch & Infer./Batch \\
\hline
Concat & 0.72M & 0.00M & 4.26s & 1.17ms \\
Grad-Blending & 0.77M & 0.05M & 13.24s & 1.20ms \\
AGM & 0.77M & 0.05M & 17.71s & 1.18ms \\
Concat+MLP & 3.04M & 2.32M & 4.82s & 1.33ms \\
Cross-Attention & 3.06M & 2.33M & 7.32s & 2.86ms \\
Sigmoid+Tanh Gate & 3.07M & 2.35M & 5.25s & 1.46ms \\
\textbf{VGMR} & \textbf{3.04M} & \textbf{2.32M} & \textbf{10.01s} & \textbf{3.87ms} \\
\hline
\end{tabular}
\end{table}

For Concat+MLP and Sigmoid+Tanh Gate, we adjust the hidden-dimensional settings to make their parameter counts close to VGMR.

As shown in Tables~\ref{tab:fusion_mosei_transformer} and~\ref{tab:mechanism_comparison_mosei}, VGMR outperforms early-fusion strategies, Cross-Attention, and generic learnable gating. 
Compared with the best early-fusion baselines on each metric, VGMR improves Accuracy and Macro-F1 by 1.48 and 1.58 percentage points, respectively. 
It also improves over Cross-Attention and Sigmoid+Tanh Gate under the same backbone setting. 
These results suggest that the gain cannot be fully attributed to stronger feature aggregation, generic cross-modal relevance modelling, or simply a larger learnable module. 
Instead, they indicate that using cross-modal agreement and discrepancy as value evidence for gate generation provides additional benefits before multimodal fusion.

Table~\ref{tab:efficiency_mosei} reports total parameters, extra parameters, training time per epoch, and inference latency per batch. 
VGMR increases the parameter count from 0.72M to 3.04M and inference latency from 1.17ms to 3.87ms per batch compared with the vanilla Transformer. 
Therefore, VGMR introduces an acceptable but non-negligible computational overhead over the simplest backbone.

However, the performance gain is unlikely to be explained solely by the increased parameter count. 
After adjusting the hidden dimensions, Concat+MLP has almost the same number of parameters as VGMR, while Sigmoid+Tanh Gate has a slightly larger parameter count than VGMR. 
Nevertheless, VGMR achieves higher Accuracy and Macro-F1 than these parameter-comparable alternatives in Tables~\ref{tab:fusion_mosei_transformer} and~\ref{tab:mechanism_comparison_mosei}. 
This suggests that the advantage of VGMR comes not only from adding more parameters, but from how these parameters are used to estimate modality value and refine features before fusion.

In terms of computational cost, VGMR is more expensive than lightweight pre-fusion alternatives, especially in inference latency. 
However, its training time remains lower than optimization-level balancing methods such as Grad-Blending and AGM, and its absolute inference latency is still within a small millisecond range in this setting. 
Overall, VGMR provides a reasonable but not cost-free performance--robustness trade-off when parameter-comparable gains and stronger noise robustness are desired.

\subsection{Compatibility with Optimization-Level Balancing}

\begin{center}
\captionof{table}{Plug-in analysis of VGMR on MOSEI with the Transformer backbone.}
\label{tab:value_gated_plugin}

\begingroup
\footnotesize
\setlength{\tabcolsep}{5pt}
\renewcommand{\arraystretch}{1.08}

\begin{tabular}{lcccc}
\hline
Method & VG Acc & VG F1 & Plain Acc & F1 \\
\hline
OGM-GE & 0.8367 (+0.0103) & 0.7914 (+0.0190) & 0.8264 & 0.7724 \\
ARL & 0.8343 (+0.0075) & 0.7870 (+0.0014) & 0.8268 & 0.7856 \\
\hline
\end{tabular}

\endgroup
\end{center}

We apply VGMR to OGM-GE and ARL under the MOSEI Transformer setting. As shown in Table~\ref{tab:value_gated_plugin}, adding VGMR consistently improves both methods. OGM-GE gains 1.01 percentage points in Accuracy and 1.90 percentage points in F1, while ARL gains 0.75 and 0.14 percentage points, respectively. These results suggest that VGMR can work together with training-stage balancing methods, since it refines modality features before fusion rather than only adjusting optimization after fusion.

\subsection{Weak-Modality Interference Analysis}

\begin{center}
\captionof{table}{Modality removal analysis on MOSEI with the Transformer backbone.}
\label{tab:mosei_modality_removal}

\begingroup
\footnotesize
\setlength{\tabcolsep}{2.8pt}
\renewcommand{\arraystretch}{1.10}

\begin{tabular}{lcccc}
\hline
Method & Full Acc & w/o Text & w/o Audio & w/o Vision \\
\hline
Concat & 0.8255 & 0.7053 (-0.1202) & 0.8092 (-0.0163) & 0.8253 (-0.0002) \\
Grad-Blending & 0.8337 & 0.7096 (-0.1241) & 0.8317 (-0.0020) & 0.8343 (+0.0006) \\
OGM-GE & 0.8264 & 0.7094 (-0.1170) & 0.8255 (-0.0009) & 0.8244 (-0.0020) \\
PMR & 0.8332 & 0.7102 (-0.1230) & 0.8324 (-0.0008) & 0.8332 (+0.0000) \\
ARL & 0.8268 & 0.7102 (-0.1166) & 0.8246 (-0.0022) & 0.8270 (+0.0002) \\
MMPareto & 0.8279 & 0.7102 (-0.1177) & 0.8289 (+0.0010) & 0.8291 (+0.0012) \\
D\&R & 0.8311 & 0.7092 (-0.1219) & 0.8302 (-0.0009) & 0.8330 (+0.0019) \\
MLA & 0.8184 & 0.7062 (-0.1122) & 0.8161 (-0.0023) & 0.8184 (+0.0000) \\
AGM & 0.8302 & 0.7105 (-0.1197) & 0.8283 (-0.0019) & 0.8296 (-0.0006) \\
\textbf{VGMR} & \textbf{0.8446} & \textbf{0.7102} (-0.1344) & \textbf{0.8410} (-0.0036) & \textbf{0.8371} (-0.0075) \\
\hline
\end{tabular}

\endgroup
\end{center}

We conduct modality removal analysis on MOSEI to examine weak-modality interference. Here, \textit{w/o Text}, \textit{w/o Audio}, and \textit{w/o Vision} denote removing one modality and using the remaining two. As shown in Table~\ref{tab:mosei_modality_removal}, removing text causes a large performance drop for all methods, confirming that text is the dominant modality. However, several balancing methods achieve higher accuracy after removing audio or vision, indicating that audio and visual modalities can introduce noisy or low-value components when directly used in joint optimization.

In contrast, VGMR’s performance decreases when either audio or vision is removed. This suggests that VGMR does not simply increase the weight of weak modalities. Instead, it uses context-conditioned value signals to guide gate generation, allowing the model to learn which weak-modality responses should be retained or reduced before fusion. Overall, the results are consistent with our motivation that the key challenge is not whether weak modalities should be used, but how to identify and retain their task-relevant parts when such parts are useful.

\subsection{Multimodal Noise Robustness Analysis}

\begin{figure}[H]
\centering
\includegraphics[width=0.95\linewidth]{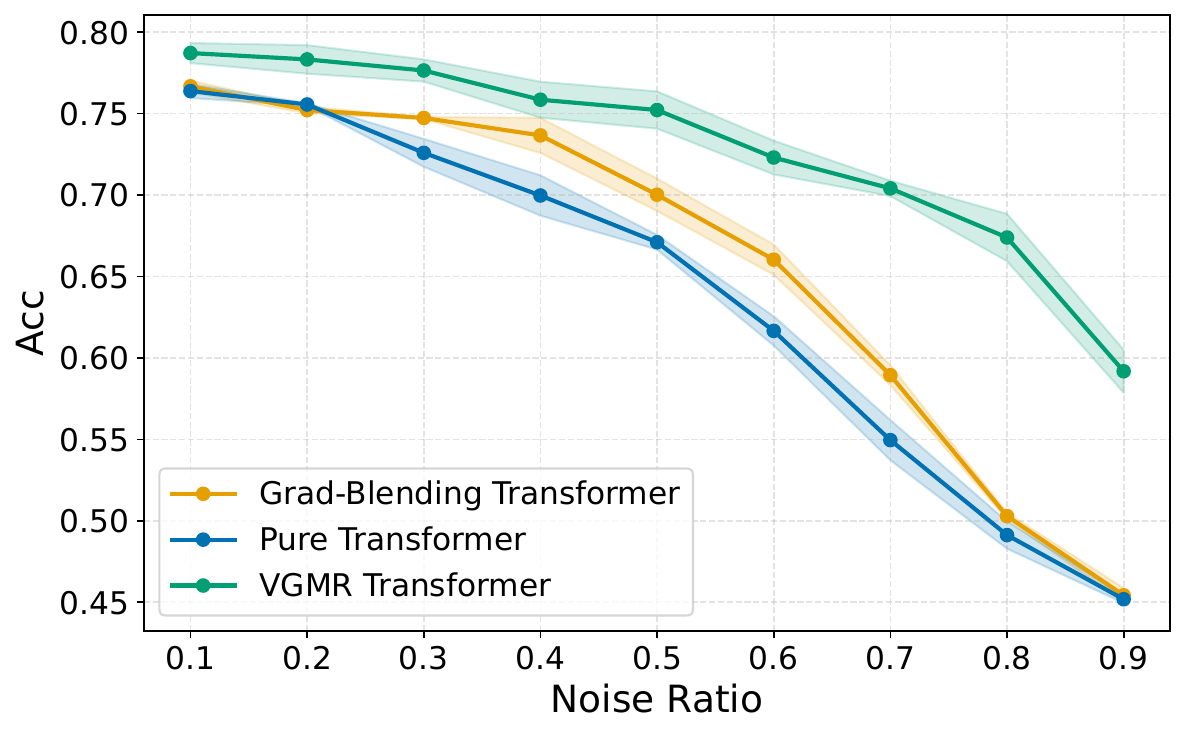}
\caption{Noise robustness on MOSI under different noise ratios.}
\label{fig:noise_ratio}
\end{figure}

\begin{figure}[!htbp]
\centering
\includegraphics[width=0.95\linewidth]{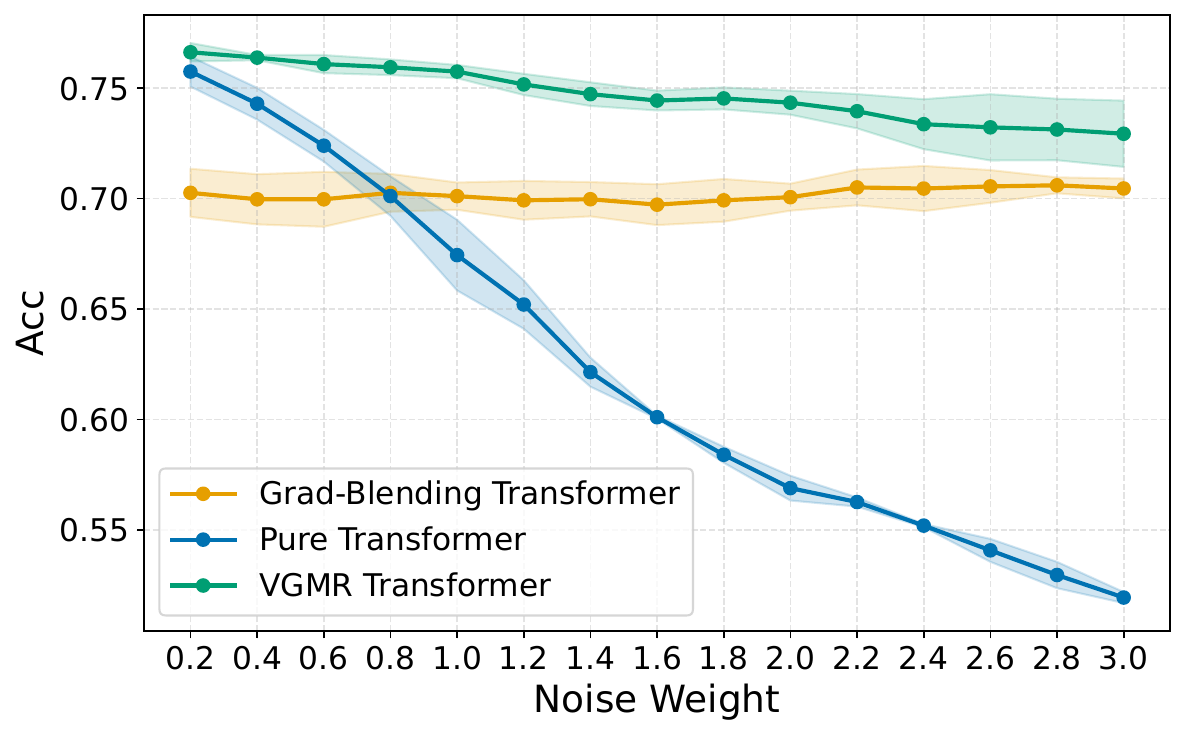}
\caption{Noise intensity robustness on MOSI under different noise weights.}
\label{fig:noise_weight}
\end{figure}

To examine robustness to corrupted modality inputs, we conduct noise experiments on MOSI by injecting Gaussian noise into all raw aligned modality features before they enter the model. Masked entries are replaced with noise sampled using the per-feature mean and standard deviation computed from the training split. We evaluate two settings: varying the noise ratio with fixed granularity, and varying the noise weight with fixed ratio. Each setting is run with three random seeds, and we report the average accuracy with variation bands.

As shown in Figures~\ref{fig:noise_ratio} and~\ref{fig:noise_weight}, all methods degrade as the noise ratio or noise weight increases, but VGMR consistently maintains higher accuracy than the Pure Transformer and Grad-Blending Transformer. The advantage becomes more visible under heavier corruption, suggesting that pre-fusion value refinement helps when multimodal inputs contain noisy components. Compared with Grad-Blending, which balances modality learning during optimization, VGMR can reduce the influence of corrupted features before fusion, leading to more stable performance under noisy multimodal conditions.

\subsection{Optimization Stability Analysis}

\begin{figure}[!htbp]
\centering
\includegraphics[width=1\linewidth]{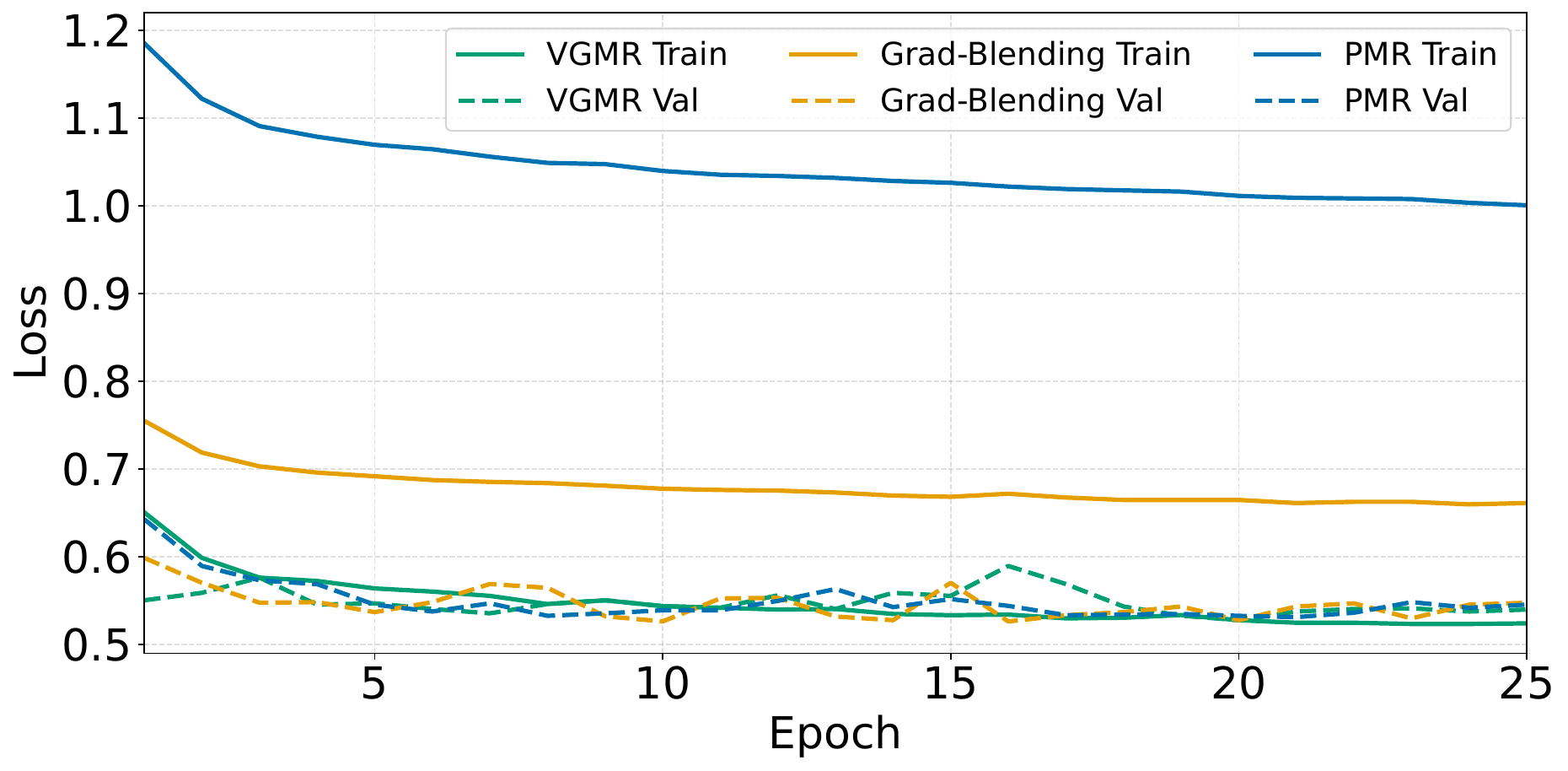}
\caption{Training and validation loss curves on MOSEI.}
\label{fig:loss_curve}
\end{figure}

We first compare the training and validation loss curves on MOSEI. As shown in Figure~\ref{fig:loss_curve}, VGMR decreases quickly at the beginning of training and maintains a smoother validation trend in later epochs. This suggests that VGMR can provide cleaner inputs to the backbone and is associated with more stable multimodal optimization.

We further examine the gradient-direction relationship between audio-only and audio-video learning, inspired by recent gradient-based analyses of modality conflict~\cite{Hao25:UniX}. For each checkpoint, we compute an audio-only gradient $g_{\text{audio}}$ and an audio-video gradient $g_{\text{audio+video}}$ on the validation split. To make them comparable, both gradients are computed with respect to the same shared parameters, including the shared Transformer backbone and final prediction head, while modality-specific input projection layers and inactive branches are excluded. The gradients are flattened and concatenated before computing:
\begin{equation}
\cos(g_{\text{audio}},g_{\text{audio+video}})
=
\frac{
g_{\text{audio}}^\top g_{\text{audio+video}}
}{
\|g_{\text{audio}}\|_2 \|g_{\text{audio+video}}\|_2
}.
\end{equation}
This metric is used as a diagnostic signal for cross-modal optimization coupling.
A large positive value indicates that the audio-video gradient is highly aligned
with the audio-only gradient, suggesting strong coupling or potential dominance
of audio-related optimization directions. A negative value indicates directional
conflict, where the joint audio-video update moves against the audio-only
optimization direction. Following UniX-style gradient diagnostics, a smoother
curve around zero is preferred, as it suggests weaker cross-modal dominance and
less severe directional conflict, while allowing relatively independent
modality contributions.

\begin{figure}[!htbp]
\centering
\includegraphics[width=0.95\linewidth]{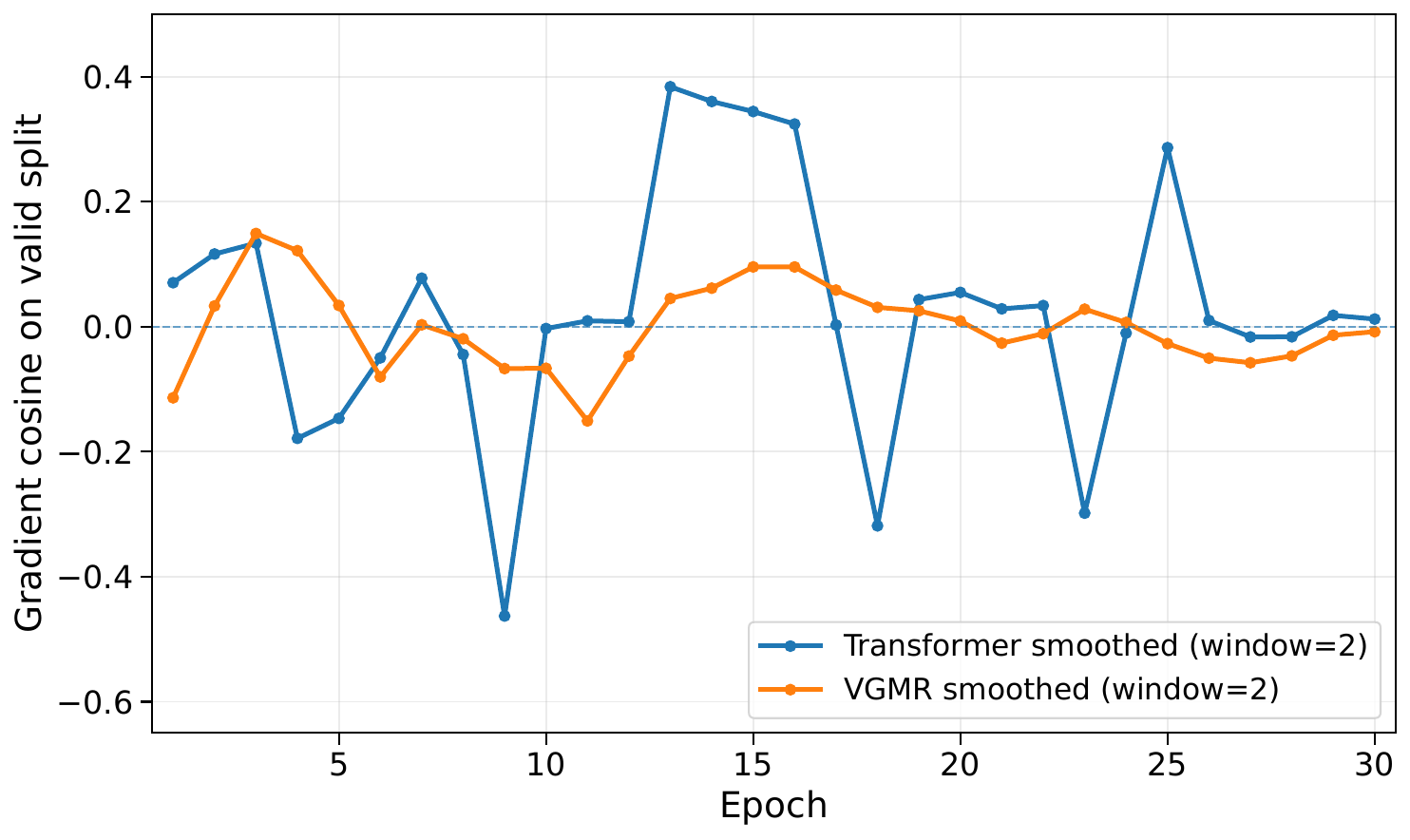}
\caption{Gradient conflict analysis on MOSEI.}
\label{fig:gradient_conflict}
\end{figure}

As shown in Figure~\ref{fig:gradient_conflict}, the plain Transformer shows larger fluctuations between positive and negative cosine values. This suggests unstable cross-modal coupling: the joint audio-video optimization direction sometimes becomes strongly aligned with the audio-only direction, but at other times moves against it. Such oscillations indicate that adding visual information may introduce unstable dominance or interference in the shared optimization
space.

In contrast, VGMR produces a smoother curve closer to zero. Following the UniX-style interpretation, this does not mean that the audio modality is ignored.
Rather, it suggests that the joint audio-video update is less dominated by a single modality-specific direction and avoids strong negative conflict. This provides diagnostic evidence consistent with our motivation that value-conditioned gate refinement can reduce unstable cross-modal coupling and encourage more balanced complementary optimization before multimodal features enter the shared backbone.

\subsection{Qualitative Case Study}

\begin{figure}[!htbp]
\centering
\includegraphics[width=0.95\linewidth]{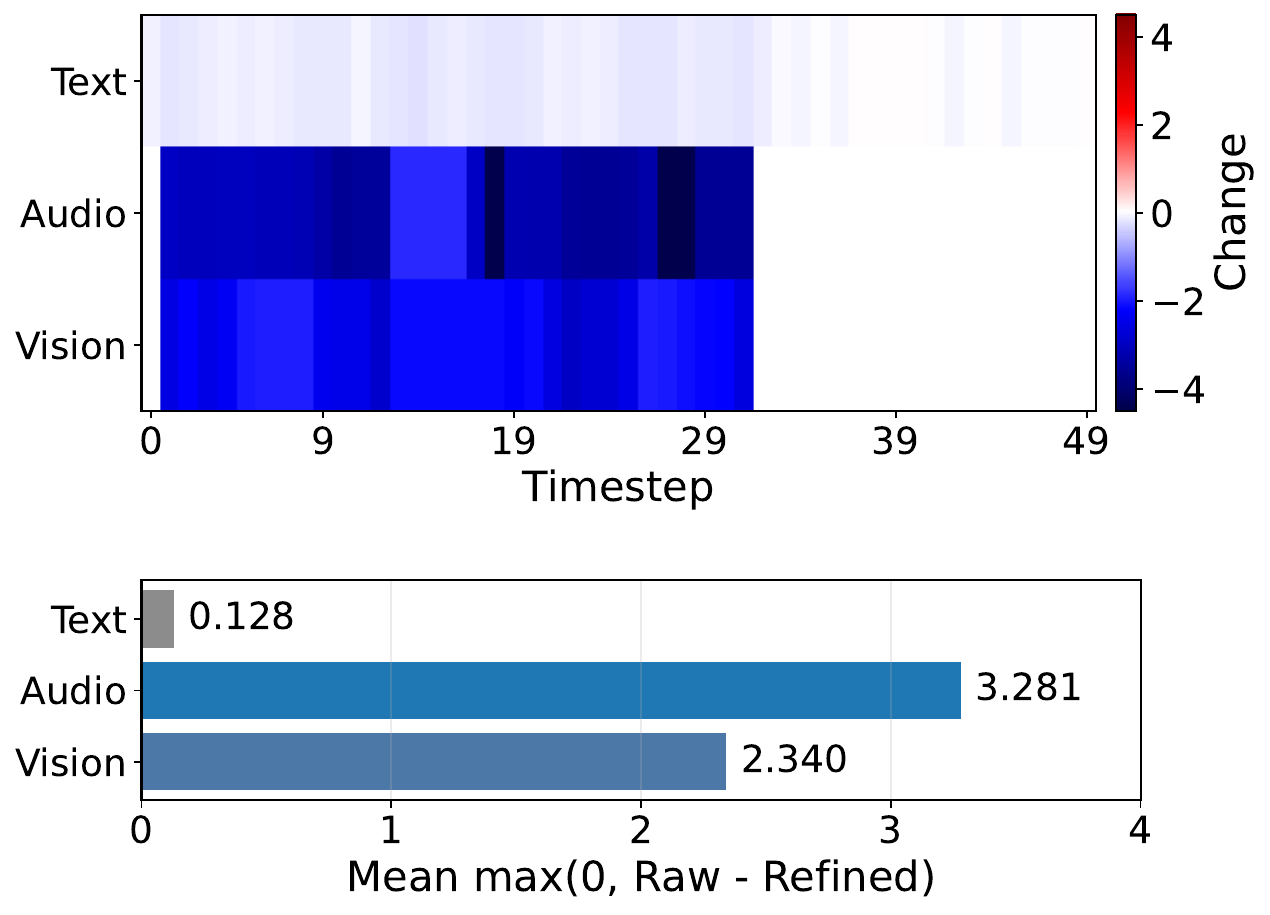}
\caption{Sample-level visualization of VGMR feature refinement on a MOSEI test case. 
Negative changes indicate suppressed feature responses.}
\label{fig:sample_change}
\end{figure}

\begin{figure}[!htbp]
\centering
\includegraphics[width=0.95\linewidth]{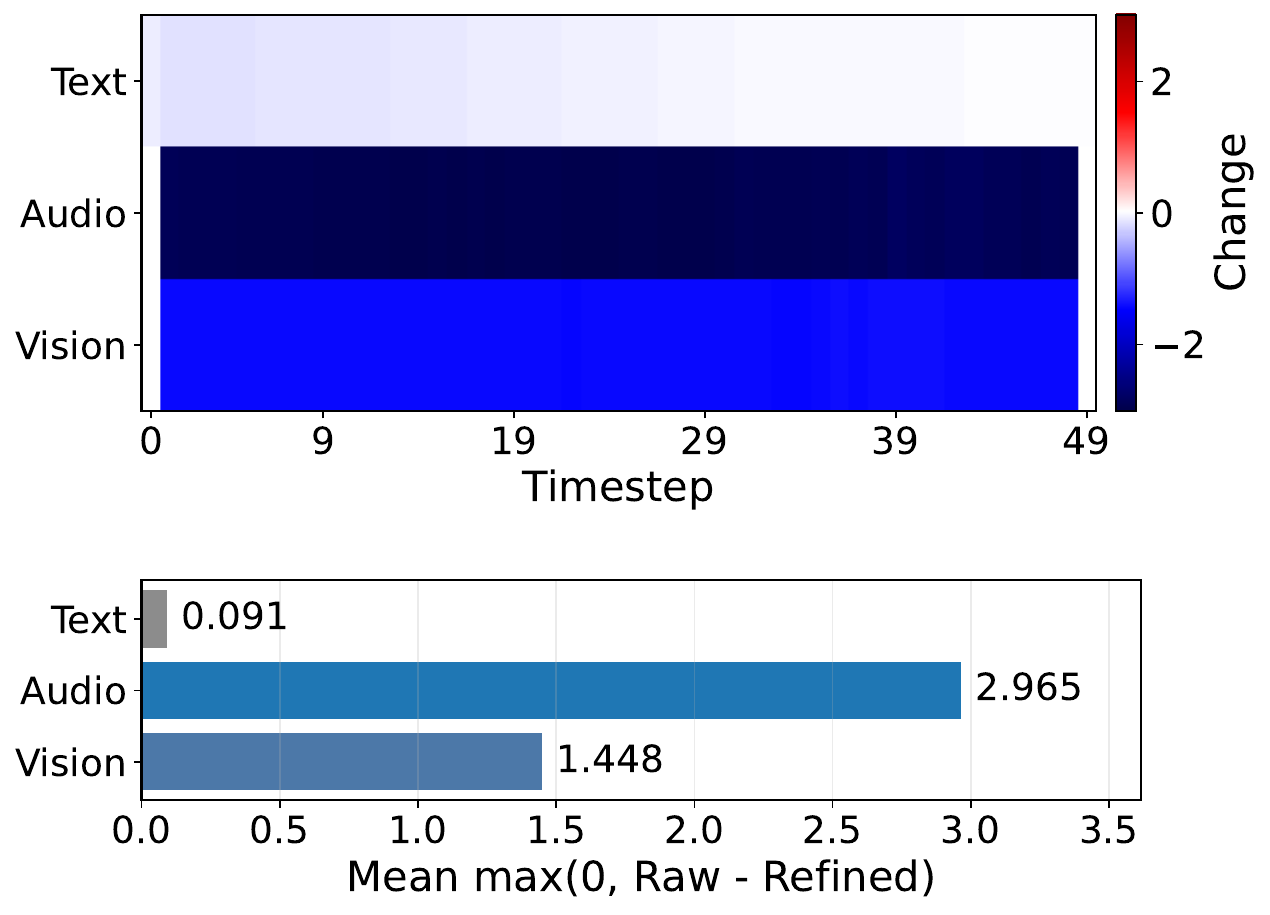}
\caption{Global visualization of average VGMR feature refinement on MOSEI. 
VGMR shows stronger suppression in audio and visual modalities than in text.}
\label{fig:global_change}
\end{figure}

To further understand how VGMR refines modality features, we present a qualitative case study on a MOSEI test sample where the baseline model gives an incorrect prediction, while VGMR produces a more accurate result. 
The ground-truth label is -2.3333. 
The baseline predicts 0.1091, whereas VGMR predicts -2.0747, which is closer to the ground truth.

As shown in Figure~\ref{fig:sample_change}, the heatmap visualizes the refinement difference 
$\Delta \mathbf{X} = \tilde{\mathbf{X}} - \mathbf{X}$, where negative values indicate suppressed feature responses. 
The bar plot reports the average suppression magnitude, computed from the positive part of $\mathbf{X} - \tilde{\mathbf{X}}$. 
For this sample, VGMR suppresses audio and visual responses more strongly, while the text modality is only slightly changed. 
This case is consistent with the view that VGMR performs instance-level and modality-aware refinement rather than applying a fixed modality-level weight.

Figure~\ref{fig:global_change} further shows the average refinement pattern on MOSEI. 
The global trend is consistent with the sample-level observation: VGMR tends to apply stronger suppression to audio and visual modalities while keeping text relatively stable.

\subsection{Summary of Ablation and Diagnostic Analyses}

Overall, the ablation and diagnostic results support the main design choices of VGMR. 
The component ablation shows that both global-level and channel-level value signals contribute to gate generation. The global value provides sample-level contextual evidence, while the channel value provides finer feature-level evidence for selective refinement. The agreement--discrepancy ablation further shows that value estimation benefits from both cross-modal consistency and conflict evidence, with discrepancy playing a particularly important role in identifying unreliable or conflicting modality responses.

The corruption and modality-removal analyses further clarify the role of the proposed value signal. Under text corruption, the estimated value scores increase while the average gate response decreases, suggesting that VGMR does not treat value as a direct preservation probability. Instead, value represents contextual impact for gate conditioning, and the gate generator combines this evidence with feature information to suppress unreliable high-impact responses. The modality-removal results also show that VGMR does not simply preserve or amplify weak modalities, but selectively retains task-relevant weak-modality information while reducing noisy or conflicting responses.

Finally, the controlled comparisons, plug-in analysis, robustness tests, gradient diagnostics, and qualitative visualizations indicate that VGMR is not merely an early-fusion transformation, a generic gating module, or a larger parameterized model. Its advantage comes from using multi-level value evidence before fusion, which can complement optimization-level balancing methods and improve robustness under noisy multimodal conditions.

\section{Future Work}

Although VGMR achieves strong or competitive performance across multiple datasets, several directions remain for future work.

First, VGMR should be further evaluated in terms of scalability and efficiency. 
As shown in Table~\ref{tab:efficiency_mosei}, it introduces acceptable but non-negligible overhead, especially in inference latency. 
Future work could test larger backbones, longer sequences, and higher-dimensional features, while exploring parameter sharing, low-rank projections, or sparse gating.

Second, value estimation can be extended from summary-level modelling to finer temporal or spatial modelling. 
The current pooled-summary design is simple and avoids strict alignment, but it may miss localized conflicts such as occlusions, frame drops, or short audio-text mismatches. 
Window-level or time-resolved value estimation may provide more detailed evidence for gate generation.

Third, the relationship between VGMR and multimodal information decomposition deserves further study. 
Although VGMR is motivated by redundancy, uniqueness, and complementarity, it does not explicitly estimate these information-theoretic quantities. 
Future work may introduce diagnostic measures or auxiliary objectives to better connect value estimation with redundancy reduction and complementary information preservation.

Finally, robustness and interpretability analysis should be expanded. 
Beyond Gaussian feature corruption and controlled text corruption, VGMR should be tested under missing modalities, temporal masking, occlusion, sensor dropouts, modality asynchrony, and distribution shifts. 
A more systematic analysis of global values, channel values, gate magnitudes, and suppression strength across modalities, classes, and noise conditions would also help characterize the learned behaviour.

\section{Conclusion}

In this paper, we revisited multimodal imbalance from a pre-fusion value estimation perspective. 
Instead of treating weak modalities as entirely useful or useless, we argued that they may contain both complementary information and harmful responses. 
Based on this motivation, we proposed Value-Gated Modality Refiner (VGMR), which estimates global-level and channel-level value evidence under cross-modal context and uses it to guide fine-grained feature refinement before fusion.

Extensive experiments and analyses show that VGMR improves multimodal learning beyond simple feature aggregation, generic gating, or training-stage rebalancing. 
The results further suggest that value-conditioned refinement can reduce unreliable modality responses before fusion, enhance robustness under noisy multimodal inputs, and support more stable optimization behaviour.

Overall, VGMR provides a pre-fusion refinement mechanism for mitigating multimodal imbalance. 
Future work will examine its scalability to larger multimodal backbones and extend the analysis to broader incomplete-modality, retrieval, recommendation, and real-world multimodal scenarios.

\section{GenAI Usage Disclosure}

Large Language Models (LLMs) were used only as writing support during the preparation of this manuscript. Their use was limited to improving grammar, wording, clarity, and overall readability. They did not contribute to the conceptual development of the work, the design of the proposed method, the experimental setup, or the interpretation of results. All research ideas, methodological decisions, analyses, and reported findings are the authors' own original work.

\bibliographystyle{ACM-Reference-Format}
\bibliography{references}
\end{document}